\documentclass[preprint,12pt]{elsarticle}

\usepackage{amssymb}
\usepackage{amsmath}
\usepackage{bm}

\usepackage{booktabs}
\newcommand{\tabincell}[2]{\begin{tabular}{@{}#1@{}}#2\end{tabular}}
\journal{Reliability Engineering and System Safety}

\begin{document}

\begin{frontmatter}



\title{Fast Collision Probability Estimation Based on Finite-Dimensional Monte Carlo Method}


\author{Zhang Hepeng$^1$, Quan Quan$^2$}

\address{1.Beijing Electro-Mechanical Engineering Institute, 40 Yungangbeili, Fengtai District, Beijing, P. R. China\\
2.Beijing University of Aeronautics and Astronautics, 37 Xueyuan Road, Haidian District, Beijing, P. R. China}

\begin{abstract}
The safety concern for unmanned systems, namely the concern for the potential casualty caused by system abnormalities, has been a bottleneck for their development, especially in populated areas. Evidently, the collision between the unmanned system and the obstacles, including both moving and static objects, accounts for a great proportion of the system abnormalities. The route planning and corresponding controller are established in order to avoid the collision, whereas, in the presence of uncertainties, it is possible that the unmanned system would deviate from the predetermined route and collide with the obstacles. Therefore, for the safety of unmanned systems, the collision probability estimation and further safety decision are very important. To estimate the collision probability, the Monte Carlo method could be applied, however, it is generally rather slow. This paper introduces a fast collision probability estimation method based on finite-dimensional distribution, whose main idea is to filter out the sampling points needed and generate the states directly by samples of finite-dimensional distribution, reducing the estimation time significantly. Besides, further techniques including the probabilistic equidistance sampling and dimension reduction, also serve to reduce the estimation time. The simulation shows that the proposed method reduces over 99\% the estimation time.
\end{abstract}


\end{frontmatter}


\section{Introduction}
The unmanned systems have been more and more popular in recent years, including unmanned aerial vehicles \cite{supervisor}, unmanned ground vehicles, autonomous ships, etc. However, the accompanying result is the various concerns, for example, the concern on environmental pollution and privacy. Among all concerns, the concern on their safety performance, namely the concern of the potential casualty caused by the unmanned system, is the first factor to consider. Evidently, the potential collision between the unmanned system and other obstacles, including both moving and static objects, accounts for a great proportion of the safety concerns.

As a matter of fact, the path planning and collision avoidance is a natural topic when it comes to the unmanned system, which has been studied for decades. Rapidly-Exploring Random Tree (RRT) \cite{RRT} was a sampling-based single-query path planning method, which operated by expanding a tree rooted at the start towards the new random sample from the closest node in the tree until it approached the goal. The velocity obstacle (VO) method \cite{VO} and reciprocal velocity obstacle (RVO) method \cite{RVO} ensured that the path was collision free given that the relative velocity was outside the VO, where RVO considered further the reaction of the obstacle. The A* algorithm \cite{A*} formed an optimal path by iterative selection of the node with the less cost in the open list, where the cost was the sum of both the cost from the start to the node and the estimated cost from the node to the goal. The artificial potential field (APF) algorithm  \cite{APF} generated a collision-free path by supposing that the agent moved under both attractive force by the goal and repulsive force by the obstacle.

Early researches on path planning archived satisfactory effects on deterministic cases. While the world is actually full of uncertainties. As for the unmanned system, the uncertainty may have three sources \cite{uncertainty}: (1)The modeling error. The mathematical model is merely an abstract of the real world, which thus could not be totally exact; (2)The observation error. The real state of the unmanned system is unknown and could only be estimated by sensors, in which process the noise is led in; (3)The disturbances on the unmanned system. For example, the acceleration caused by wind.

In the presence of uncertainty, the unmanned system may deviate from its predetermined route and collide with the obstacle. That is why the uncertainty was incorporated in many of the recent researches. The Chance Constrained RRT (CC-RRT) algorithm \cite{CC-RRT} was an improvement of the RRT algorithm, which ensured the probabilistic feasibility of the planned path by bounding the collision probability (CP) at each time step within a certain range.
In \cite{LQG-MP}, a number of candidate paths were first generated by the RRT algorithm, then the best one among them was selected with respect to the chosen planning objective based on the a-priori probability distribution of the system state. The simulation showed that the quality of the candidate paths differed starkly with the uncertainty. The Probabilistic RVO (PRVO) method \cite{PRVO} improved the RVO method to assure that the CP is less than a certain threshold utilizing the Bayesian decomposition. In \cite{B-UAVC}, the probabilistic collision avoidance was guaranteed, namely the inter-robot CP was below a user-specified threshold, by constraining each robot's motion within a certain separated area called buffered uncertainty-aware Voronoi cell (B-UAVC). \cite{PrSBC} guaranteed the probabilistic safety if a certain constraint was imposed on the input. \cite{PU} managed to improve the robustness of the model predictive control collision avoidance method for the autonomous ground vehicle with respect to the parametric uncertainties.

These researches did take the uncertainty into consideration, while the CP, as the key measurement of safety, was not detailed. They at most gave a sufficient condition that the CP was within a certain range. For better evaluation of the safety under uncertainty, the CP should be estimated precisely. In this paper, the CP is defined mathematically as the probability that the unmanned system collides with an obstacle or another unmanned system in a certain time interval to come.

To estimate the CP, the Monte Carlo (MC) method \cite{MCbook} could be utilized, which is to estimate the CP with the collision frequency, after a number of independent trials. The MC method is relatively accurate, while it generally takes a lot of time to accomplish the estimation. Therefore, a fast estimation method is needed. In \cite{condition}, a collision estimation method was proposed based on conditional probability. The whole route was discretized to a number of sampling points, and a conditional probability that the collision did not occur at the current sampling point given that the collision did not occur for all previous sampling points was estimated. The CP was estimated by multiplying all the conditional probabilities. However, the CP estimation may tend to 1 with the increment of sampling points in some cases. Moreover, the method was suitable mainly for the 2-dimensional collision. In \cite{PCE}, the CP was estimated based on polynomial chaos expansion. Its main idea was to approximate the solution of the system model with polynomial chaos expansion, based on which the MC method was applied to estimate the CP. However, this method applied mainly to random variables instead of random processes. In \cite{NASA}, the CP was estimated with an integral within a certain area in the case of free flight. However, the assumption that the relative velocity and error ellipse was constant made it not precise enough. Moreover, it applied mainly to the case of collision with only one obstacle. In \cite{MDTC}-\cite{BBN}, the ship-ship CP was estimated with the product of the potential encounter numbers and the probability that both ships failed to evade the collision. Whereas, it was rather an overall CP than that of a specific case. For example, in \cite{SouthKorea}, the CP was estimated for a scenario of maritime spent nuclear fuel transportation. The ship knew little about the predetermined route on departure: whether it would encounter another ship, where the potential encounters would occur, and which type another ship would be. The CP was estimated based on experience and historical data collected from the Automatic Identification System (AIS). Therefore, it could not estimate the CP for a specific case precisely.

In this paper, we come up with a Finite-Dimensional Monte Carlo (FDMC) method, which attempts to reduce the estimation time of the MC method. The MC method consumes a lot of time, mainly due to three disadvantages. First, in the MC method, a number of propagations are needed in the state generation, and dense sampling points are required to ensure precision. However, collision generally occurs on few sampling points. While, we have no idea neither on which sampling points the collide may occur, nor how could the state of a certain sampling point be generated directly without a lot of time-consuming propagations. Second, the MC method propagates the whole state, which may include the velocity and attitude, etc. While it is merely the position that should be taken into consideration when it comes to the collision problem. Third, the MC method selects the sampling points in an equitime manner. However, there should be more sampling points where the velocity is greater and less where the velocity is smaller. Four improvements are led in in this paper to eliminate these disadvantages. First, the state is proved to be a normal process, which makes it possible to be generated directly. Second, the importance sampling filters out those sampling points that are prone to collision. These two improvements eliminate the first disadvantage -- the sampling points with a great CP could now be filtered out and generated directly. In this procedure, the states are generated directly with samples of a finite-dimensional distribution (FDD) instead of state propagation in the MC method. Third, the dimension reduction generates directly the position, which eliminates the second disadvantage. Last, the probabilistic equidistant sampling determines the density of the sampling points proportional to the velocity, eliminating the third disadvantage. With all these improvements, the CP estimation time has been reduced greatly, which is verified by the simulation.

The remainder of this paper is organized as follow. Some preliminaries are given in section II. The problem is formulated in section III. The FDMC method is presented in section IV. The simulation follows in section V, and the paper ends with the conclusion in section VI.
\section{Preliminaries}
For better clarification of our method, first, some relevant definitions and theorems are led in.

\textbf{Definition 1}(Probability Space): The probability space is defined as the triplet $(\Omega,\mathcal{F},\Pr)$, where $\Omega$ is the sample space, $\mathcal{F}\subset2^\Omega$ is a non-empty subset of the power set of $\Omega$, and the probability $\Pr:\mathcal{F}\rightarrow [0,1]$ maps each element (defined as an event) $A\in\mathcal{F}$ to a real number $\Pr(A)\in[0,1]$, with $\Pr(\Omega)=1,\Pr(\varnothing)=0$,
\begin{equation}
\Pr\left(\bigcup_{k=1}^\infty A_k\right)=\sum_{k=1}^\infty \Pr(A_k),\forall A_k\in\mathcal{F},A_k\cap A_l=\varnothing,k\ne l.
\end{equation}

\textbf{Definition 2}(Normal Distribution): An $n$-dimensional random vector $\mathbf{X}$ is said to obey the \textit{Normal Distribution}(denoted as $\mathbf{X}\sim\mathcal{N}(\bm \mu,\bm \Sigma)$), if it has the probability density function(pdf)
\begin{equation}
f_{\mathbf{X}}(\mathbf{x})=\frac{1}{\sqrt{2\pi}^n|\bm\Sigma|^{\frac{1}{2}}}\exp\left(-\frac{(\mathbf{x}-\bm\mu^{\text{T}})\bm\Sigma^{-1}(\mathbf{x}-\bm\mu)}{2}\right)
\end{equation}
where $\bm\mu=\mathbb{E}[\mathbf{X}]$ is the expectation vector, and $\bm\Sigma=\mathbf{Cov}[\mathbf{X},\mathbf{X}]$ is the covariance matrix, which is positive definite. The $\alpha$-quantile $z_\alpha$ is defined for the standard normal distribution $X\sim\mathcal{N}(0,1)$ such that $\Pr(X>z_\alpha)=\alpha$.

\iffalse
\textbf{Definition 3}(Normal Process): A random process $\mathbf{X}(t)$ is said to be a \textbf{Normal Process} if any FDD of $\mathbf{X}(t)$ is a normal distribution, i.e., $\mathbf{Y}=[\begin{array}{cccc}\mathbf{X}^{\text{T}}(t_1) & \mathbf{X}^{\text{T}}(t_2) & \cdots & \mathbf{X}^{\text{T}}(t_n) \end{array}]^{\text{T}}\sim\mathcal{N}(\bm \mu,\bm \Sigma)$ is normally distributed, $\forall t_1,t_2,\cdots,t_n$.
\fi

\textbf{Definition 3}(Normal Process): A random process $\mathbf{X}(t)$ is said to be a \textit{Normal Process} if any FDD of $\mathbf{X}(t)$ is a normal distribution, i.e., $\forall t_1,t_2,\cdots,$\\
$t_n,\mathbf{Y}=[\begin{matrix}\mathbf{X}^{\text{T}}(t_1) & \mathbf{X}^{\text{T}}(t_2) & \cdots & \mathbf{X}^{\text{T}}(t_n) \end{matrix}]^{\text{T}}\sim\mathcal{N}(\bm \mu,\bm \Sigma)$ is normally distributed.

\textbf{Definition 4}(Brownian Motion): The normal process $\mathbf{B}(t)$ is said to be \textit{Standard Brownian Motion}, if
1)$\mathbf{B}(0)=\mathbf{0}$;
2)the increment is standard and stationary , i.e., $\forall s,t>0,\mathbf{B}(s+t)-\mathbf{B}(s)\sim\mathcal{N}(\mathbf{0},t\mathbf{I})$;
3)the increments are independent, i.e., $\forall 0\le t_0<t_1<t_2$, the increments $\mathbf{B}(t_2)-\mathbf{B}(t_1)$ and $\mathbf{B}(t_1)-\mathbf{B}(t_0)$ are independent.

\textbf{Theorem 1}: Suppose that $\mathbf{x}(t)$ is the solution to the equation
\begin{equation}
\text{d}\mathbf{x}(t)=\mathbf{A}\mathbf{x}(t)\text{d}t+\mathbf{c}\text{d}t+\bm\Sigma\text{d}\mathbf{B}(t),\mathbf{x}(0)\sim\mathcal{N}(\bm \mu_0,\bm \Sigma_0)
\end{equation}
where $\mathbf{c}$ is a constant vector, and $\mathbf{A},\bm\Sigma$ are constant matrices of proper size,
then $\mathbf{x}(t)$ is a normal process with the expectation and covariance
\begin{align}
\mathbb{E}\left[\mathbf{x}(t)\right]=&e^{\mathbf{A}t}(\bm\mu_0+\int_0^te^{-\mathbf{A}\tau}\mathbf{c}\text{d}\tau)\notag\\
\mathbf{Cov}\left[\mathbf{x}(s),\mathbf{x}(t)\right]=&e^{\mathbf{A}s}\left(\bm\Sigma_0+\int_0^{\min(s,t)} e^{-\mathbf{A}u}\bm\Sigma\bm\Sigma^{\text{T}}e^{-\mathbf{A}^{\text{T}}u}\text{d}u\right)e^{\mathbf{A}^\text{T}t}.
\end{align}

The proof is given in the appendix.

\textbf{Theorem 2}: Suppose that $x(t)$ is the solution to
\begin{equation}
\text{d}\dot{x}+k_\text{d}\dot{x}\text{d}t+k_\text{p}x\text{d}t
=c\text{d}t+G\text{d}B(t)\label{dynamic_x}
\end{equation}
with initial state normally distributed
\begin{equation}
\left[\begin{matrix}x(0)\\\dot{x}(0)\end{matrix}\right]
\sim\mathcal{N}\left(
\left[\begin{matrix}\mu_0\\\dot{\mu}_0\end{matrix}\right]
,
\left[\begin{matrix}\sigma_x^2 & \rho\sigma_x\sigma_{\dot{x}}\\ \rho\sigma_x\sigma_{\dot{x}} & \sigma_{\dot{x}}^2\end{matrix}\right]
\right).
\end{equation}
Here, $k_\text{d},k_\text{p},c,G$ are constants, and $B(t)$ is the 1D standard Brownian motion. If the corresponding characteristic equation
\begin{equation}
\lambda^2+k_\text{d}\lambda+k_\text{p}=0
\end{equation}
has two unequal real roots $\lambda_1\ne\lambda_2,\lambda_1,\lambda_2\ne0$, then $x(t)$ is a normal process with the expectation
\begin{align*}
\mathbb{E}[x(t)]&=\frac{\lambda_2e^{\lambda_1t}-\lambda_1e^{\lambda_2t}}{\lambda_2-\lambda_1}\mu_0
+
\frac{e^{\lambda_2t}-e^{\lambda_1t}}{\lambda_2-\lambda_1}\dot{\mu}_0
+c\frac{\lambda_1e^{\lambda_2t}-\lambda_2e^{\lambda_1t}+\lambda_2-\lambda_1}{k_\text{p}(\lambda_2-\lambda_1)}
\end{align*}
and covariance
\begin{align*}
\text{Cov}[x(s),x(t)]&=C_x+C_{\dot{x}}+C_{x,\dot{x}}+C_u
\end{align*}
where $s<t$ and
\begin{footnotesize}
\begin{align*}
C_x&=\frac{(\lambda_2e^{\lambda_1s}-\lambda_1e^{\lambda_2s})(\lambda_2e^{\lambda_1t}-\lambda_1e^{\lambda_2t})}{(\lambda_2-\lambda_1)^2}\sigma_x^2\\
C_{\dot{x}}&=\frac{(e^{\lambda_2s}-e^{\lambda_1s})(e^{\lambda_2t}-e^{\lambda_1t})}{(\lambda_2-\lambda_1)^2}\sigma_{\dot{x}}^2\\
C_{x,\dot{x}}&=\frac{(\lambda_2e^{\lambda_1s}-\lambda_1e^{\lambda_2s})(e^{\lambda_2t}-e^{\lambda_1t})+(\lambda_2e^{\lambda_1t}-\lambda_1e^{\lambda_2t})(e^{\lambda_2s}-e^{\lambda_1s})}{(\lambda_2-\lambda_1)^2}\rho\sigma_x\sigma_{\dot{x}}\\
C_u&=G^2\left(\frac{\lambda_1e^{\lambda_2(t-s)}-\lambda_2e^{\lambda_1(t-s)}}{-2k_\text{p}k_\text{d}(\lambda_2-\lambda_1)}
+\frac{\lambda_1e^{\lambda_2(s+t)}
+\lambda_2e^{\lambda_1(s+t)}   }
{2k_\text{p}(\lambda_2-\lambda_1)^2}+
\frac{
e^{\lambda_2s+\lambda_1t}
+e^{\lambda_2t+\lambda_1s}
}
{k_\text{d}(\lambda_2-\lambda_1)^2}\right).
\end{align*}
\end{footnotesize}
Proof:

The solution to (\ref{dynamic_x}) is
\begin{align*}
x(t)&=\frac{\lambda_2e^{\lambda_1t}-\lambda_1e^{\lambda_2t}}{\lambda_2-\lambda_1}x(0)
+
\frac{e^{\lambda_2t}-e^{\lambda_1t}}{\lambda_2-\lambda_1}\dot{x}(0)
+c\frac{\lambda_1e^{\lambda_2t}-\lambda_2e^{\lambda_1t}+\lambda_2-\lambda_1}{k_\text{p}(\lambda_2-\lambda_1)}\\
&+e^{\lambda_1t}\int_0^te^{(\lambda_2-\lambda_1)v}\text{d}v\int_0^{v}e^{-\lambda_2u}G\text{d}B(u)
\end{align*}
and the calculation of the expectation and covariance is similar to that of \textit{Theorem 1}, which is thus omitted here.$\square$

\textbf{Definition 5}(Chi-Square Distribution): Suppose that $n$ random variables are independently and identically distributed $X_1,X_2,\cdots,X_n\text{ i.i.d}\sim\mathcal{N}(0,1)$, then $\mathcal{X}^2=\sum_{i=1}^nX_i^2$ is a chi-square distribution with degree of freedom (DOF) $n$, denoted as $\mathcal{X}^2\sim\mathcal{X}^2(n)$. The $\alpha$-quantile $\mathcal{X}^2_\alpha(n)$ is defined such that $\Pr(\mathcal{X}^2>\mathcal{X}^2_\alpha(n))=\alpha$.

\textbf{Proposition 1}: Suppose that $\mathbf{X}\sim\mathcal{N}(\bm \mu,\bm \Sigma)$ with dimension $n$, then $(\mathbf{X}-\bm\mu)^{\text{T}}\bm\Sigma^{-1}(\mathbf{X}-\bm\mu)$ is chi-squared distributed with DOF $n$
\begin{equation}
(\mathbf{X}-\bm\mu)^{\text{T}}\bm\Sigma^{-1}(\mathbf{X}-\bm\mu)\sim\mathcal{X}^2(n).
\end{equation}

Proof:

Notice that the positive definite covariance matrix $\bm\Sigma$ could be decomposed as $\bm\Sigma=\mathbf{C}^\text{T}\mathbf{C}$, let $\mathbf{X}^\prime={\mathbf{C}^\text{T}}^{-1}(\mathbf{X}-\bm\mu)$, then $(\mathbf{X}-\bm\mu)^{\text{T}}\bm\Sigma^{-1}(\mathbf{X}-\bm\mu)={\mathbf{X}^\prime}^\text{T}\mathbf{X}^\prime$ with $\mathbf{X}^\prime\sim\mathcal{N}(\mathbf{0},\mathbf{I})$. The conclusion is drawn directly by the definition of the chi-square distribution.$\square$

\begin{figure}[tbh]
\centering\includegraphics[scale=0.3]{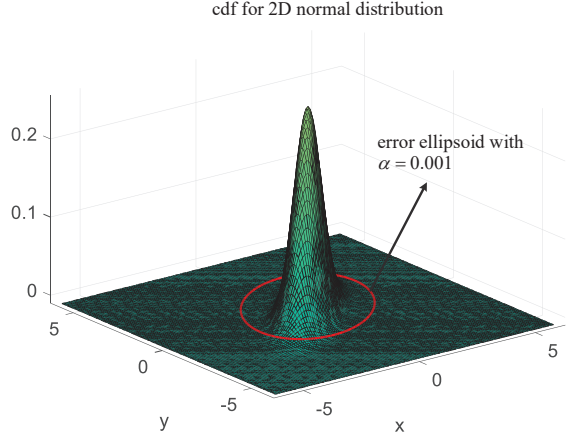}
\caption{Error Ellipsoid}
\label{ErrorEllipsoid}
\end{figure}

\textbf{Definition 6}: The ellipsoid
\begin{equation}
E_\alpha:(\mathbf{X}-\bm\mu)^{\text{T}}\bm\Sigma^{-1}(\mathbf{X}-\bm\mu)\le\mathcal{X}^2_\alpha(n)\label{ellipseun}
\end{equation}
is called the error ellipsoid of the $n$-dimensional normal distribution $\mathbf{X}\sim\mathcal{N}(\bm \mu,\bm \Sigma)$ with confidence $1-\alpha$. It implies that $\mathbf{X}$ lies in $E_\alpha$ with probability $1-\alpha$
\begin{equation}
\Pr(\mathbf{X}\in E_\alpha)=\iint_{E_\alpha}f_{\mathbf{X}}(\mathbf{x})\text{d}\mathbf{x}=1-\alpha
\end{equation}
as is shown in Fig. \ref{ErrorEllipsoid}.

\textbf{Definition 7}(Population): A population is a set of individuals $P=\left\{I_i,i=1,2,\cdots,N\right\}$ with each individual attached to a certain fitness value $f(I_i)$, where the fitness function is defined as $f:P\rightarrow [0,\infty)$. The cumulative frequency is defined as
\begin{equation}
CF_i=\frac{\sum_{j=1}^if(I_j)}{\sum_{j=1}^Nf(I_j)}.
\end{equation}

\textbf{Theorem 3}(Roulette Wheel Selection): With the aforementioned population $P$,
let the selection be $S=I_i$ in the case that $CF_{i-1}\le R<CF_i$, where $R\sim U[0,1]$ is uniformly distributed, then the probability that $I_i$ is selected is proportional to its fitness $f(I_i)$
\begin{equation}
\Pr(S=I_i)=\frac{f(I_i)}{\sum_{j=1}^Nf(I_j)}\propto f(I_i)
\end{equation}
with $\sum_{i=1}^N\Pr(S=I_i)=1$.

The conclusion is evident and is omitted here.

\textbf{Remark 1}: In the situation where $I_i$ are continuous quantities, and the repeated selections are not expected, the selection $S$ could be modified to
\begin{equation}
S=\frac{I_i+I_{i-1}}{2}+\frac{R-CF_{i-1}}{CF_i-CF_{i-1}}\frac{I_{i+1}-I_{i-1}}{2}.
\end{equation}

\begin{figure}[tbh]
\centering\includegraphics[scale=0.5]{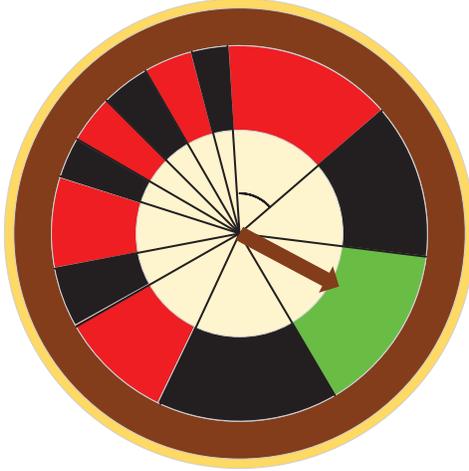}
\caption{Roulette Wheel Selection}
\label{Roulette}
\end{figure}

\section{Modeling and Problem Formulation}
\subsection{CP Modelling}
Consider the linear dynamic model with uncertainty of the unmanned system $S$,
\begin{equation}
\text{d}\mathbf{x}(t)=\mathbf{A}\mathbf{x}(t)\text{d}t+\mathbf{c}\text{d}t+\bm\Sigma\text{d}\mathbf{B}(t),\mathbf{x}(0)\sim\mathcal{N}(\bm \mu_0,\bm \Sigma_0)\label{dynamic}
\end{equation}
where $\mathbf{x}=[\begin{matrix}\mathbf{p}^\text{T} & \mathbf{v}^\text{T}\end{matrix}]^{\text{T}}$ with $\mathbf{p}=[\begin{matrix}x & y & z\end{matrix}]^{\text{T}}$ the position of the barycenter and $\mathbf{v}=\dot{\mathbf{p}}$ the velocity, and there exist transform matrices $\mathbf{T}_{\mathbf{p}},\mathbf{T}_{\mathbf{v}}$ such that $\mathbf{p}=\mathbf{T}_{\mathbf{p}}\mathbf{x},\mathbf{v}=\mathbf{T}_{\mathbf{v}}\mathbf{x}$; $\mathbf{A}$ is the system matrix, $\mathbf{c}$ is a constant vector, $\mathbf{B}(t)$ is the uncertainty in the form of the standard Brownian Motion, which is independent with the initial state $\mathbf{x}(0)$, and $\bm\Sigma$ is the matrix which represents the properties of the uncertainty.

In the case where the channels $x,y,z$ are independent, (\ref{dynamic}) could be divided into three parts, where the dynamic model of the $x$ channel could be rewritten as
\begin{equation}
\text{d}\dot{x}+k_\text{d}\dot{x}\text{d}t+k_\text{p}x\text{d}t
=c\text{d}t+G\text{d}B(t)
\end{equation}
with initial state normally distributed
\begin{equation}
\left[\begin{matrix}x(0)\\\dot{x}(0)\end{matrix}\right]
\sim\mathcal{N}\left(
\left[\begin{matrix}\mu_0\\\dot{\mu}_0\end{matrix}\right]
,
\left[\begin{matrix}\sigma_x^2 & \rho\sigma_x\sigma_{\dot{x}}\\ \rho\sigma_x\sigma_{\dot{x}} & \sigma_{\dot{x}}^2\end{matrix}\right]
\right).
\end{equation}
Here, $k_\text{p},k_\text{d}$ are control parameters, $c$ is constant, $B(t)$ is the 1D standard Brownian motion, and $G$ represents the property of the uncertainty. The $y,z$ channels are similar and omitted here.

Suppose that there exist $k_\text{O}$ obstacles $O_j,j=1,2,\cdots,k_\text{O}$, which, together with the unmanned system $S$, are all considered to be 3D rigid bodies in the form of a sphere
\begin{align*}
S(t)&=\left\{\mathbf{p}\left|\parallel\mathbf{p}-\mathbf{p}(t)\parallel_2\le r\right.\right\}\\
O_j(t)&=\left\{\mathbf{p}\left|\parallel\mathbf{p}-\mathbf{o}_j(t)\parallel_2\le R_j\right.\right\}
\end{align*}
where $\mathbf{p}(t)$ and $r$ are the barycenter and radius of $S$, $\mathbf{o}_j(t)$ and $R_j$ are the geometric center and radius of the obstacle $O_j$, and $\parallel\cdot\parallel_2$ is the Euclidean norm of a vector.

To formulate the CP, first the probability space $(\Omega,\mathcal{F},\Pr)$ should be defined, with the sample space $\Omega$ defined as
\begin{equation}
\Omega=\left\{\mathbf{x}(t)|\text{d}\mathbf{x}(t)=\mathbf{A}\mathbf{x}(t)\text{d}t+\mathbf{c}\text{d}t+\bm\Sigma\text{d}\mathbf{B}(t),\mathbf{x}(0)\sim\mathcal{N}(\bm \mu_{0},\bm \Sigma_{0})\right\}
\end{equation}
namely the set of all the paths that are subject to the dynamic model (\ref{dynamic}).
The collision event on $[0,T]$ is defined
\begin{equation}
C_{[0,T]}=\left\{\mathbf{x}(t)|\exists t\in\left[0,T\right],\exists j=1,2,\cdots,k_\text{O},S(t)\bigcap O_j(t)\neq \varnothing\right\}
\end{equation}
as the event that the positions of the unmanned system $S$ and the obstacle $O_j$ intersect. The CP $\Pr(C_{[0,T]})$ could then be defined.

The exact value of the CP $\Pr(C_{[0,T]})$ is not available; as a result, it should be estimated. The MC estimation could be utilized in this procedure.
\subsection{CP Estimation Based on the MC Method}
The main idea of the MC method is to propagate the dynamic model (\ref{dynamic}) to calculate the states at different stages and judge whether the collision occurs. After a number of independent trials, the CP is estimated with the collision frequency.

In this procedure, there are three kinds of probabilities.
\begin{itemize}
\item Firstly, the real CP $\Pr(C_{[0,T]})$. It is the probability defined on a continuous path, which has been defined in the previous subsection.
\item Secondly, the discrete CP $\Pr(C_{[0,T],N})$. The continuous path should be discretized to a number of sampling points that are dense enough to represent the original path. Similar with $C_{[0,T]}$, the collision event in the discrete form could be defined as
    \begin{small}
    \begin{equation}
C_{[0,T],N}=\left\{\mathbf{x}(t)|\exists k=1,2,\cdots,N,\exists j=1,2,\cdots,k_\text{O},S(t_k)\bigcap O_j(t_k)\neq \varnothing\right\}
\end{equation}
\end{small}
where $t_k=k\tau$ is the $k$th sampling point out of $N$, and $\tau=\frac{T}{N}$ is the sampling step. The discrete CP is utilized to approximate the true CP $\Pr(C_{[0,T]})$, provided that the sampling points are dense enough.

\item Finally, the discrete collision frequency Fr$(C_{[0,T],N},M)$ with $M$ independent trials. The discrete CP $\Pr(C_{[0,T],N})$ is generally difficult to calculate directly, therefore, the MC method estimates it with the corresponding frequency Fr$(C_{[0,T],N},M)$, defined as
    \begin{equation}
    \text{Fr}(C_{[0,T],N},M)=\frac{1}{M}\sum_{i=1}^{M}I_i(C_{[0,T],N})
    \end{equation}
    where $I_i(C_{[0,T],N}),i=1,2,\cdots,M$ are a set of independent and identically distributed samples from the Bernoulli distribution $I(C_{[0,T],N})\sim \text{Ber}(\Pr(C_{[0,T],N}))$, which takes value 1 if $C_{[0,T],N}$ occurs and 0 otherwise. In each sample, the state is propagated from $t_k$ to $t_{k+1}$ through the discrete form of (\ref{dynamic})
    \begin{equation}
    \mathbf{x}(t_{k+1})=(\mathbf{I}+\mathbf{A}\tau)\mathbf{x}(t_k)+\tau\mathbf{c}+\bm\Sigma(\mathbf{B}(t_{k+1})-\mathbf{B}(t_k)).
    \end{equation}
\end{itemize}
With the following \textit{Theorem 4}, the CP $\Pr(C_{[0,T]})$ could be estimated with $\text{Fr}(C_{[0,T],N},M)$.

\textbf{Theorem 4}: $\text{Fr}(C_{[0,T],N},M)$ converges towards $\Pr(C_{[0,T]})$ with probability 1
\begin{equation}
\Pr\left(\lim_{M,N\rightarrow \infty}\text{Fr}(C_{[0,T],N},M)=\Pr(C_{[0,T]})\right)=1
\end{equation}
with approximate $100\alpha\%$ confidence that the estimation error
\begin{small}
\begin{equation}
\left|\Pr(C_{[0,T]})-\text{Fr}(C_{[0,T],N},M)\right|\le z_{\alpha/2}\sqrt{\frac{1}{M}\text{Fr}(C_{[0,T],N},M)\left(1-\text{Fr}(C_{[0,T],N},M)\right)}.
\end{equation}
\end{small}

The proof is given in the appendix.

The procedure is summarised in Fig. \ref{MC}. As we can see, the flow includes the inner loop and the outer loop, where the inner loop generates the states $\mathbf{x}(t_1),\cdots,\mathbf{x}(t_N)$ along the path and judge whether the collision occurs in this single trial. In other words, it serves to generate the sample $I_i(C_{[0,T],N})$. While the outer loop counts the total times of collision $c$ and further estimates the CP $\hat{\Pr}(C_{[0,T]})$ with the collision frequency $\frac{c}{M}$ for the $M$ independent trials. It is evident that the estimation time is proportional to both $M$ the number of independent trials and $N$ the number of sampling points, therefore, the time complexity of the MC method is
\begin{equation}
T(M,N)=O(MN).
\end{equation}
\begin{figure}[tbh]
\centering\includegraphics[scale=0.75]{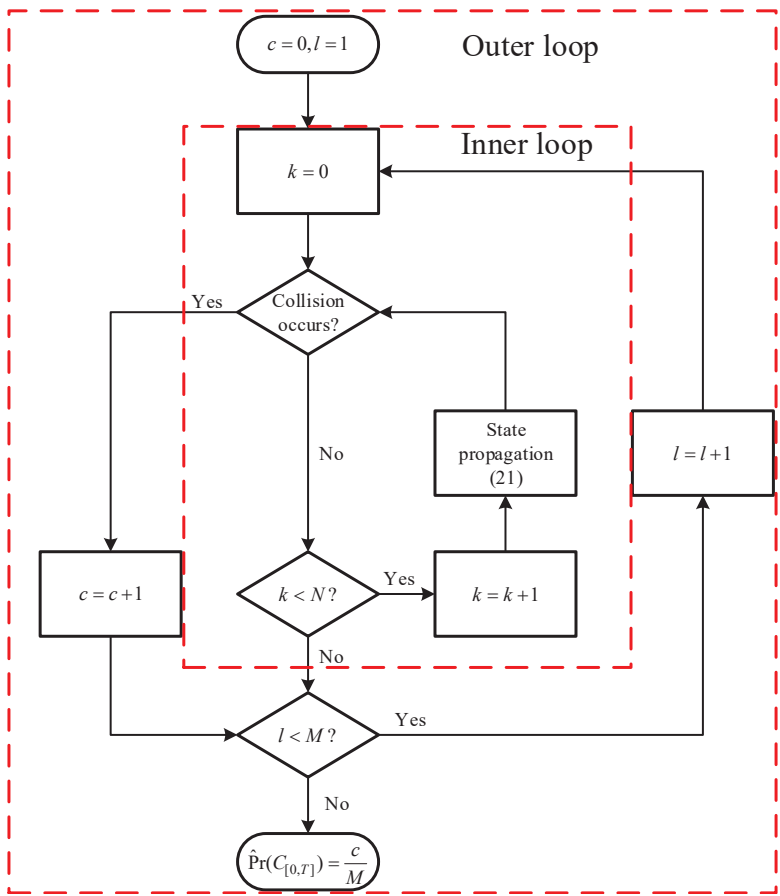}
\caption{Flow Chart of the MC Method}
\label{MC}
\end{figure}
\subsection{Problem Formulation}
With the MC method, the CP could be estimated by the collision frequency. However, as is known to us, the MC method generally consumes a lot of time. Hence, our problem is formulated as:

\textbf{CP Estimation Problem}: For the unmanned system $S$ subject to the dynamic model (\ref{dynamic}) and the given time interval $\left[0,T\right]$, the CP $\Pr(C_{[0,T]})$ could be estimated as $\hat\Pr(C_{[0,T]})=\text{Fr}(C_{[0,T],N},M)$ by the time-consuming MC method according to \textit{Theorem 4}. Then how could an alternative method be proposed, which is equivalent with the MC method to the estimation time reduced greatly?

\section{CP Estimation Based on the FDMC Method}
In this section, the FDMC method is presented in three aspects. Subsection 4.1 clarifies the theory of the FDMC method, subsection 4.2 supplies details and improvements, while subsection 4.3 ends this section with the summary and discussion of the FDMC method.
\subsection{Basic Theory}
\textit{Theorem 4} indicates that, the number of independent trials $M$ and the number of sampling points $N$ in one trial must be great enough to ensure the precision in the MC method, which is the cause of the long estimation time. The idea of the FDMC method is to ensure the precision with a much smaller amount of sampling points $n\ll N$ to reduce the estimation time.

As is shown in Fig. \ref{MC}, the MC method is implemented through the state propagation. While it has an obvious disadvantage. In practice, the collision may only occur within few states that are near the obstacle. However, to obtain these states, a number of state propagations are necessary, which is evidently a time-consuming procedure. Here, an alternative method is introduced to replace the state propagation of the MC method, which considers the direct generation of states that are near the obstacle. This procedure is done by \textit{Theorem 1} or \textit{Theorem 2}.

\textit{Theorem 1} indicates that the solution $\mathbf{x}(t)$ to the dynamic model (\ref{dynamic}) is a normal process. Therefore, any FDD of $\mathbf{x}(t)$ is normally distributed, namely,
\begin{equation}
\mathbf{X}=\left[\begin{matrix}\mathbf{x}^{\text{T}}(s_1) & \cdots & \mathbf{x}^{\text{T}}(s_n)\end{matrix}\right]^{\text{T}}\sim\mathcal{N}(\bm \mu_\mathbf{X},\bm \Sigma_\mathbf{X}),\forall n>0,s_1,\cdots,s_n\in[0,T]
\end{equation}
with
\begin{equation}
\bm\mu_{\mathbf{X}}=
\left[\begin{matrix}\mathbb{E}\left[\mathbf{x}(s_1)\right] \\ \vdots \\ \mathbb{E}\left[\mathbf{x}(s_n)\right]\end{matrix}\right],
\bm \Sigma_{\mathbf{X}}=
\left[
\begin{matrix}
\mathbf{Cov}\left[\mathbf{x}(s_1),\mathbf{x}(s_1)\right] & \cdots & \mathbf{Cov}\left[\mathbf{x}(s_1),\mathbf{x}(s_n)\right]\\
\vdots & \ddots & \vdots\\
\mathbf{Cov}\left[\mathbf{x}(s_n),\mathbf{x}(s_1)\right] & \cdots & \mathbf{Cov}\left[\mathbf{x}(s_n),\mathbf{x}(s_n)\right]
\end{matrix}
\right].
\end{equation}
In the case where $x,y,z$ channels are independent, it is more convenient to reduce the dimension by calculating the FDD of the position $\mathbf{P}=[\begin{matrix}x^{\text{T}}(s_1) & \cdots & x^{\text{T}}(s_n)\end{matrix}]^{\text{T}}\sim\mathcal{N}(\bm \mu_\mathbf{P},\bm \Sigma_\mathbf{P})$ directly through \textit{Theorem 2}.

\begin{figure}[tbh]
\centering\includegraphics[scale=0.65]{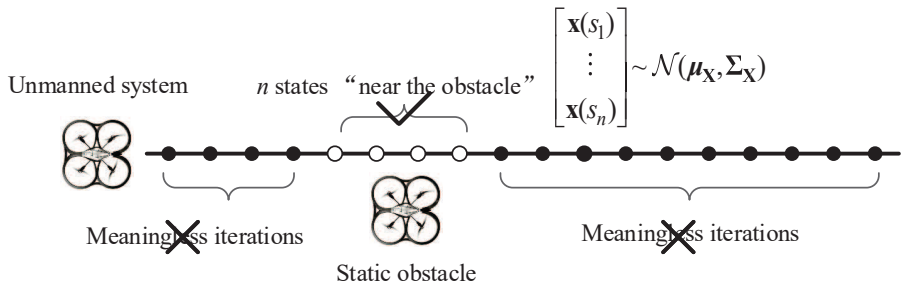}
\caption{Generation of the States near the obstacle through Finite-Dimensional Distribution}
\label{comparison}
\end{figure}

Therefore, it is possible to select $n\ll N$ sampling points out of $N$ that are near the obstacles and generate them with a normal distribution instead of the state propagation, reducing a great deal of estimation time, as is shown in Fig. \ref{comparison}. Only these states are to be considered in the CP estimation.
Similarly, there are also two kinds of probabilities.
\begin{itemize}
\item Firstly, the discrete CP $\Pr(C_{[s_1,s_n],n})$, where the event $C_{[s_1,s_n],n}$ is defined as
\begin{small}
    \begin{equation}
C_{[s_1,s_n],n}
=\left\{\mathbf{x}(t)|\exists k=1,2,\cdots,n,\exists j=1,2,\cdots,k_\text{O},S(s_k)\bigcap O_j(s_k)\neq \varnothing\right\}.
\end{equation}
\end{small}

\item Secondly, the discrete collision frequency Fr$(C_{[s_1,s_n],n},M)$ with $M$ independent trials, defined as
    \begin{equation}
    \text{Fr}(C_{[s_1,s_n],n},M)=\frac{1}{M}\sum_{i=1}^{M}I_i(C_{[s_1,s_n],n})
    \end{equation}
    where $I_i(C_{[s_1,s_n],n}),i=1,2,\cdots,M$ are a set of independently and identically distributed samples from the Bernoulli distribution $I(C_{[s_1,s_n],n})$\\$\sim \text{Ber}(\Pr(C_{[s_1,s_n],n}))$, which takes value 1 if $C_{[s_1,s_n],n}$ occurs and 0 otherwise. In each sample, the states are generated directly through the normal distribution $\mathbf{X}\sim\mathcal{N}(\bm \mu_\mathbf{X},\bm \Sigma_\mathbf{X})$.
\end{itemize}

\textbf{Theorem 5}: It is assumed that $\left|\Pr(C_{[s_1,s_n],n})-\Pr(C_{[0,T]})\right|\le\varepsilon$ with $\exists \varepsilon>0$ a small constant, then we have approximate $100\alpha\%$ confidence that the estimation error
\begin{footnotesize}
\begin{equation}
\left|\Pr(C_{[0,T]})-\text{Fr}(C_{[s_1,s_n],n},M)\right|\le z_{\alpha/2}\sqrt{\frac{1}{M}\text{Fr}(C_{[s_1,s_n],n},M)\left(1-\text{Fr}(C_{[s_1,s_n],n},M)\right)}+\varepsilon.
\end{equation}
\end{footnotesize}

The proof is similar with that of \textit{Theorem 4}, which is thus omitted here.

\textbf{Remark 2}: In \textit{Theorem 5}, an assumption is made that the distance between $\Pr(C_{[s_1,s_n],n})$ and $\Pr(C_{[0,T]})$ is small enough. As a matter of fact, on the one hand, $C_{[s_1,s_n],n}$ occurs $\Rightarrow C_{[0,T]}$ occurs; on the other hand, given that $C_{[0,T]}$ occurs, it is almost guaranteed that $C_{[s_1,s_n],n}$ occurs, as $s_1,\cdots,s_n$ are the nearest sampling points from the obstacles. In other words, the sampling points $s_1,\cdots,s_n$ are \textit{representative}. Therefore, $\Pr(C_{[s_1,s_n],n})$ is close to $\Pr(C_{[0,T]})$, and the assumption is reasonable.

\begin{figure}[tbh]
\centering\includegraphics[scale=0.75]{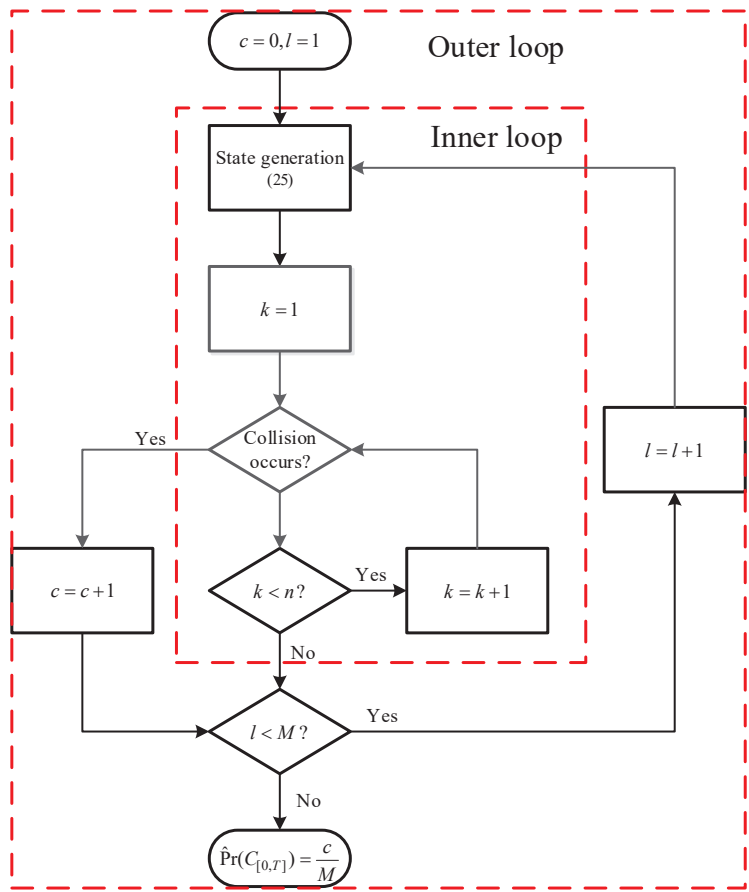}
\caption{Flow Chart of FDMC Method}
\label{FDMC}
\end{figure}

The procedure is summarised in Fig. \ref{FDMC}. It is shown that the inner loop is simplified by replacing $N$ times the state propagation with a sample of the FDD, while the $M$ independent trials in the outer loop are unchanged. In other words, compared with \textit{Theorem 4} of the MC method, FDMC method only requires that $M$ is great enough, namely it reduces the time complexity $T(M,N)=O(MN)$ of the MC method to $T(M)=O(M)$, which of course reduces a great deal of the estimation time. In the meanwhile, there is little loss of the precision, as $\varepsilon$ is a small constant.

Only the improvement on the number $N$ of the sampling points are made in this paper, and the improvement on the number $M$ of the independent trials are to be studied in future researches.
\subsection{Details and improvements}
The basic theory of the FDMC method is presented in the previous subsection, whereas, there remain two questions to answer: 1) how is the term "near the obstacle" defined, namely how could the $n$ sampling points $s_1,\cdots,s_n$ be determined? 2)in which form are the sampling points organized? This subsection completes the FDMC method by answering these two questions.
\subsubsection{Importance sampling}
The importance sampling answers the first question by defining mathematically the term "near the obstacle" as the points where the CP $\Pr(S\cap O_j\ne\varnothing)$ is high enough to consider. It is a filtering procedure which filters out the sampling points where the CP is high enough and ignores the others, which is carried out by \textit{Theorem 6}.

\textbf{Theorem 6}: Recall that the position $\mathbf{p}(s_k)$ of $S$ is normally distributed at the moment $s_k$ with radius $r$, and the center of the $j$th obstacle $O_j$ is $\mathbf{o}_j(s_k)$, with radius $R_j$. $E_\alpha$ is the error ellipsoid of $S$ with confidence $1-\alpha$. Define the augmented obstacle sphere $O_{\text{a},j}=\left\{\mathbf{p}|\parallel\mathbf{p}-\mathbf{o}_j(s_k)\parallel_2\le r+R_j\right\}$, which includes the radii of both $S$ and $O_j$. Then a sufficient condition that the CP between $S$ and $O_j$ is less than $\alpha$ is that $O_{\text{a},j}$ does not intersect with $E_{\alpha}$, namely
\begin{equation}
O_{\text{a},j}\cap E_{\alpha}=\varnothing\Rightarrow\Pr(S\cap O_j\neq \varnothing)<\alpha.
\end{equation}

Proof:

Notice that the collision occurs if and only if the barycenter of $S$ lies in $O_{\text{a},j}$, namely $\mathbf{p}(s_k)\in O_{\text{a},j}\Leftrightarrow S\cap O_j\neq \varnothing$, then
\begin{footnotesize}
\begin{equation}
\Pr(S\cap O_j\neq\varnothing)=\Pr(\mathbf{p}(s_k)\in O_{\text{a},j})
=\iint_{O_{\text{a},j}}f_{\mathbf{p}(s_k)}(\mathbf{p})\text{d}\mathbf{p}<1-\iint_{E_\alpha}f_{\mathbf{p}(s_k)}(\mathbf{p})\text{d}\mathbf{p}=\alpha
\end{equation}
\end{footnotesize}
with $f_{\mathbf{p}(s_k)}(\mathbf{p})$ pdf of $\mathbf{p}(s_k)$.
$\square$

\begin{figure}[tbh]
\centering\includegraphics[scale=0.75]{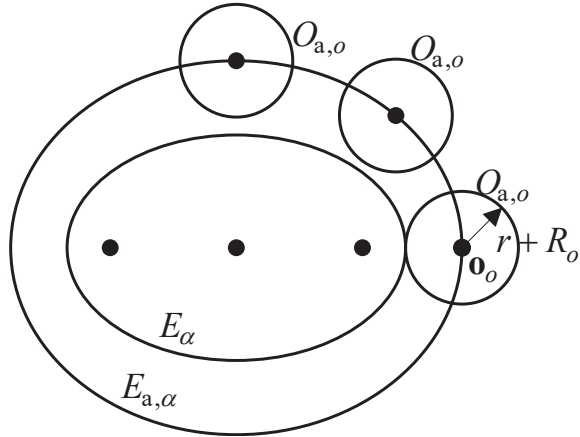}
\caption{A cross-section of the error ellipsoid}
\label{Th2}
\end{figure}

With \textit{Theorem 6}, the nearest sampling points $s_1,\cdots,s_n$ from the obstacles could be determined. There remains the last question: in which form are the sampling points organized?
\subsubsection{Equidistant Sampling in Probability}
In the MC method, the equitime sampling is usually applied. While it has two disadvantages. First, when the velocity of the unmanned system is rather high, it is highly likely that the discrete collision event $C_{[0,T],N}$ would omit collision moment between sampling points; second, when the velocity is rather low, the sampling points are rather dense, and the estimation time is wasted. Therefore, the second question is answered with the equidistant sampling. Notice that the solution $\mathbf{x}(t)$ to (\ref{dynamic}) is given by \textit{Theorem 1}, then the probabilistic equidistant sampling is to determine a set of $N_{\text{ed}}$ sampling points $t_{\text{ed},k},k=1,2,\cdots,N_{\text{ed}}$ such that their expected positions are equidistant
\begin{equation}
\mathbb{E}[\mathbf{p}(t_{\text{ed},k})]=\mathbf{T}_{\mathbf{p}}\mathbb{E}[\mathbf{x}(t_{\text{ed},k})]=\mathbf{p}_{\text{ed},k}\label{equidistance}
\end{equation}
where $\mathbf{p}_{\text{ed},k}=[\begin{matrix}x_{\text{ed},k} & y_{\text{ed},k} & z_{\text{ed},k}\end{matrix}]^{\text{T}}$ are a set of equidistant positions. In the case where $x,y,z$ channels are independent, the dimension could be reduced by replacing (\ref{equidistance}) with the much simpler 1D equation
\begin{equation}
\mathbb{E}[x(t_{\text{ed},k})]=x_{\text{ed},k}.\label{equidistance1D}
\end{equation}
While, sometimes, it may be not easy to solve equation (\ref{equidistance}) or (\ref{equidistance1D}), thus an alternative method is given.

Recall that the sampling points in the MC method are selected in an equitime manner
\begin{equation}
t_k=k\tau,k=1,2,\cdots,N.
\end{equation}
By \textit{Theorem 1}, the expected states $\mathbb{E}\left[\mathbf{x}(t_k)\right]$ could be calculated, based on which the expected velocities $\mathbb{E}\left[\mathbf{v}(t_k)\right]=\mathbf{T}_{\mathbf{v}}\mathbb{E}\left[\mathbf{x}(t_k)\right]$ could be obtained. Then $t_{\text{ed},k}$ could be obtained by $N_{\text{ed}}$ independent rounds of the roulette wheel selection by \textit{Theorem 3}, where the individuals are the equitime sampling points $t_k$ with the expected velocity $\parallel\mathbb{E}\left[\mathbf{v}(t_k)\right]\parallel_2$ as the fitness.

\subsection{Summary and Discussion}
Up to now, the \textit{CP Estimation Problem} has been solved, the flow of which has been summarised in Table \ref{flow}.

\begin{table}[!hbp]
\caption{FDMC Flow}%
\centering \
\begin{tabular}
[c]{l|l}%
\toprule
Step 1 & \tabincell{l}{Selection of a set of $N_\text{ed}$ equidistant sampling points\\ $t_{\text{ed},k}\in[0,T],k=1,2,\cdots,N_\text{ed}$, by the roulette wheel selection\\ of \textit{Theorem 3} or solving equation (\ref{equidistance}) or (\ref{equidistance1D});}\\
\hline
Step 2 & \tabincell{l}{Determination of $n\ll N_\text{ed}$ sampling points $s_k,k=1,2,\cdots,n$\\out of $t_{\text{ed},k}$,which are near the obstacle, according to \textit{Theorem 6};}\\
\hline
Step 3 & \tabincell{l}{Determination of the FDD $\mathbf{P}\sim\mathcal{N}(\bm \mu_\mathbf{P},\bm \Sigma_\mathbf{P})$, namely \\calculation of $\bm \mu_\mathbf{P}$ and $\bm \Sigma_\mathbf{P}$, by \textit{Theorem 1} or \textit{Theorem 2};}\\
\hline
Step 4 & \tabincell{l}{Estimation of the CP $\Pr(C_{[0,T]})$ with the collision frequency\\$\text{Fr}(C_{[s_1,s_n],n},M)$, according to \textit{Theorem 5}.}\\
\bottomrule
\end{tabular}\label{flow}
\end{table}

Compared with the MC method, the FDMC method could reduce a great deal of the estimation time due to the three reasons
\begin{itemize}
\item The importance sampling, which could eliminate most of the low-collision-probability sampling points;
\item The dimension reduction, which could eliminate other parts of the state except the position.
\item The probabilistic equidistant sampling, which could reduce the amount of sampling points of low velocity;
\end{itemize}

\section{Simulation}
\begin{figure}[tbh]
\centering\includegraphics[scale=0.5]{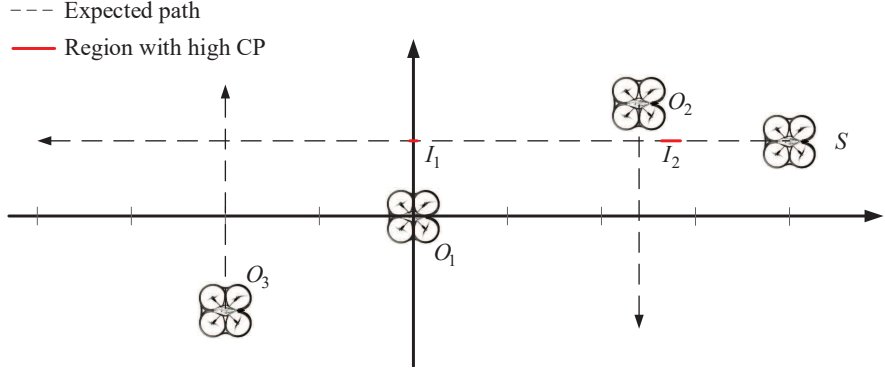}
\caption{Simulation Scene}
\label{simulationScene}
\end{figure}

\begin{figure}[tbh]
\centering\includegraphics[scale=0.5]{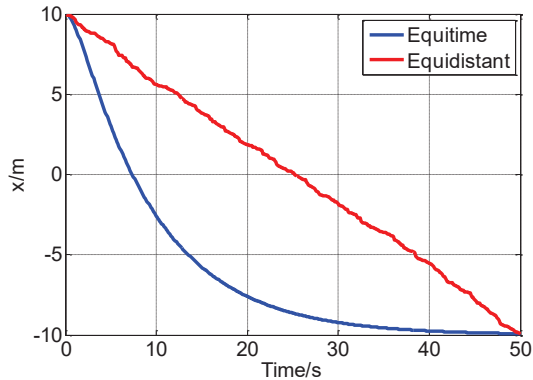}
\caption{Variation of $x$ position of sampling points with time in both manners}
\label{equidistant}
\end{figure}

The quadcopter $S$ is controlled to fly in the negative direction of the $x$-axis, starting from $[10\text{m},2\text{m}]$. Along the path, there exist three other quadcopters $O_1,O_2,O_3$, as is shown in Fig. \ref{simulationScene}.
In this case, it is $S$ that is in our control, and our target is to estimate the probability that $S$ collides with $O_1,O_2,O_3$ for some time to come. Therefore, the quadcopters $O_1,O_2,O_3$ are deemed the obstacles, where $O_1$ is hovering at the origin, and $O_2,O_3$ are moving along straight lines. All these quadcopters are moving within the same horizontal plane. Some other parameters are shown in Table \ref{sp}.
\begin{table}[ptbh]
\caption{Simulation Parameters}%
\label{sp}%
\centering \
\begin{tabular}
[c]{l|l}%
\toprule
Quadcopter radii $r,R_1,R_2,R_3$ & $0.75\text{m}$\\
Starting point of $O_2$ & $[6\text{m},3\text{m}]^\text{T}$\\
Starting point of $O_3$ & $[-5\text{m},-3\text{m}]^\text{T}$\\
Velocity of $O_2,O_3$ & 1m$\cdot$s$^{-1}$\\
Time interval $[0,T]$ & $[0,50\text{s}]$\\
Trial times $M$ & $10^6$\\
\bottomrule
\end{tabular}
\end{table}
The simulation in MATLAB R2014a is carried out automatically, whereas, for better clarification, the flow is analyzed step by step. First, by Step 1, a set of probabilistic equidistant sampling points are selected and compared with the equitime ones. Fig. \ref{equidistant} shows how the expected $x$ position varies over time in both sampling manners.
It could be inferred that the quadcopter $S$ decelerates over time. It is not advisable to sample with the same density in $[0,10\text{s}]$ and $[40\text{s},50\text{s}]$, the velocity of which is evidently much smaller than the former one. As a matter of fact, sampling points in $[40\text{s},50\text{s}]$ is merely a waste of the estimation time, which is avoided thanks to the probabilistic equidistant sampling.

Second, the importance sampling of Step 2 is carried out, as is shown in Fig. \ref{simulationScene}. It is indicated that the collision may only occur on few moments: probable collision with $O_1$ on $I_1$ and $O_2$ on $I_2$. As for the rest, they are not to be considered in Step 3 and 4. There is no doubt that the importance sampling would reduce the estimation time significantly.

After the calculation of Step 3 and 4, the CP estimation result is shown in Table \ref{timeComparison}.
\begin{table}[ptbh]
\caption{Comparison of the estimation time}%
\label{timeComparison}%
\centering \
\begin{tabular}
[c]{l|ll}%
\toprule
& CP estimation & estimation time\\
\midrule
FDMC & 0.004997$\pm$0.000138 & 95.792s\\
MC   & 0.005461$\pm$0.000144 & 20471.655s\\
\bottomrule
\end{tabular}
\end{table}
It is shown that the FDMC method reduces 99.53\% the estimation time compared with the MC method with little loss of precision, which has verified the effectiveness of the FDMC method in the case of CP estimation with multiple static or moving obstacles.
\section{Conclusion}
With the development of the unmanned system, there have been more and more concerns on their safety performance, among which the collision problem is drawing more and more attention. In this paper, we come up with an FDMC method, which is much faster than the MC method in the CP estimation. First, those moments with high CP are filtered out by the importance sampling, and the states are generated directly by a sample of the normal process, without the time-consuming propagations. Second, the dimension reduction makes it possible to generate merely the position, with the rest of the state ignored. Third, the probabilistic equidistant sampling reduces the number of sampling points with low CP. With these improvements, the estimation time has been reduced greatly. The simulation proves the effectiveness of the proposed method, which reduces over 99\% the estimation time.

Actually, there remain a lot of aspects to improve. The model studied in this paper is the linear time-invariant system, which should be extended to the systems that are more general, namely linear time-variant systems and nonlinear systems. The determination of the parameter of the uncertainty is not yet mentioned in this paper, and the method is to be modified to be data-driven. The method proposed in this paper, though improved a lot compared with the MC method, remains an MC method in nature. It could be more analytical, if the CP could be calculated by an integral of the pdf. In that case, the CP could be estimated absolutely in real time. In all, there remains a lot of work to do.




\appendix

\subsection{Proof of Theorem 1}
\textbf{Theorem 1}: Suppose that $\mathbf{x}(t)$ is the solution to the equation
\begin{equation}
\text{d}\mathbf{x}(t)=\mathbf{A}\mathbf{x}(t)\text{d}t+\mathbf{c}\text{d}t+\bm\Sigma\text{d}\mathbf{B}(t),\mathbf{x}(0)\sim\mathcal{N}(\bm \mu_0,\bm \Sigma_0)
\end{equation}
where $\mathbf{c}$ is a constant vector, and $\mathbf{A},\bm\Sigma$ are constant matrices of proper size,
then $\mathbf{x}(t)$ is a normal process with the expectation and covariance
\begin{align}
\mathbb{E}\left[\mathbf{x}(t)\right]=&e^{\mathbf{A}t}(\bm\mu_0+\int_0^te^{-\mathbf{A}\tau}\mathbf{c}\text{d}\tau)\notag\\
\mathbf{Cov}\left[\mathbf{x}(s),\mathbf{x}(t)\right]=&e^{\mathbf{A}s}\left(\bm\Sigma_0+\int_0^{\min(s,t)} e^{-\mathbf{A}u}\bm\Sigma\bm\Sigma^{\text{T}}e^{-\mathbf{A}^{\text{T}}u}\text{d}u\right)e^{\mathbf{A}^\text{T}t}.
\end{align}

Proof:

The solution
\begin{equation}
\mathbf{x}(t)=e^{\mathbf{A}t}\mathbf{x}(0)+e^{\mathbf{A}t}\int_0^te^{-\mathbf{A}\tau}\mathbf{c}\text{d}\tau+e^{\mathbf{A}t}\int_0^te^{-\mathbf{A}\tau}\bm\Sigma\text{d}\mathbf{B}(\tau)
\end{equation}
is a normal process because
\begin{equation}
\int_0^te^{-\mathbf{A}\tau}\bm\Sigma\text{d}\mathbf{B}(\tau)=\lim_{n\rightarrow\infty}\frac{t}{n}\sum_{i=1}^ne^{-\mathbf{A}\frac{it}{n}}\bm\Sigma\text{d}\mathbf{B}(\frac{it}{n})
\end{equation}
as a linear combination of normal processes, is a normal process. For the expectation $\mathbb{E}\left[\mathbf{x}(t)\right]$ and covariance $\mathbf{Cov}\left[\mathbf{x}(s),\mathbf{x}(t)\right]$, notice that
the Brownian motion has a zero mean
\begin{equation}
\mathbb{E}\left[\text{d}\mathbf{B}(u)\right]=\mathbf{0}
\end{equation}
and independent increment
\begin{equation}
\mathbb{E}\left[\text{d}\mathbf{B}(u)\text{d}\mathbf{B}^{\text{T}}(v)\right]=
\left\{
\begin{matrix}
\mathbf{I}\text{d}u & u=v\\
\mathbf{0} & u\ne v
\end{matrix}
\right.
\end{equation}
therefore,
\begin{equation}
\mathbb{E}\left[\int_0^te^{-\mathbf{A}\tau}\bm\Sigma\text{d}\mathbf{B}(\tau)\right]=\int_0^te^{-\mathbf{A}\tau}\bm\Sigma\mathbb{E}\left[\text{d}\mathbf{B}(\tau)\right]=\mathbf{0}
\end{equation}
and
\begin{footnotesize}
\begin{align*}
\mathbb{E}\left[\int_0^se^{-\mathbf{A}\tau}\bm\Sigma\text{d}\mathbf{B}(\tau)\int_0^t\text{d}\mathbf{B}^{\text{T}}(\tau)\bm\Sigma^{\text{T}}e^{-\mathbf{A}^{\text{T}}\tau}\right]
&=\iint e^{-\mathbf{A}u}\bm\Sigma\mathbb{E}[\text{d}\mathbf{B}(u)\text{d}\mathbf{B}^{\text{T}}(v)]\bm\Sigma^{\text{T}}e^{-\mathbf{A}^{\text{T}}v}\\
&=\int_0^{\min(s,t)}e^{-\mathbf{A}u}\bm\Sigma\bm\Sigma^{\text{T}}e^{-\mathbf{A}^{\text{T}}u}\text{d}u.
\end{align*}
\end{footnotesize}
Finally the expectation is obtained, and the covariance could be calculated by
\begin{equation}
\mathbf{Cov}\left[\mathbf{x}(s),\mathbf{x}(t)\right]=\mathbb{E}\left[\mathbf{x}(s)\mathbf{x}^{\text{T}}(t)\right]-\mathbb{E}\left[\mathbf{x}(s)\right]\mathbb{E}\left[\mathbf{x}^{\text{T}}(t)\right].
\end{equation}
$\square$
\subsection{Proof of Theorem 4}
\textbf{Theorem 4}: $\text{Fr}(C_{[0,T],N},M)$ converges towards $\Pr(C_{[0,T]})$ with probability 1
\begin{equation}
\Pr\left(\lim_{M,N\rightarrow \infty}\text{Fr}(C_{[0,T],N},M)=\Pr(C_{[0,T]})\right)=1
\end{equation}
with approximate $100\alpha\%$ confidence that the estimation error
\begin{small}
\begin{equation}
\left|\Pr(C_{[0,T]})-\text{Fr}(C_{[0,T],N},M)\right|\le z_{\alpha/2}\sqrt{\frac{1}{M}\text{Fr}(C_{[0,T],N},M)\left(1-\text{Fr}(C_{[0,T],N},M)\right)}.
\end{equation}
\end{small}

Proof:

Evidently
\begin{equation}
\lim_{N\rightarrow \infty}\Pr(C_{[0,T],N})=\Pr(C_{[0,T]})
\end{equation}
and by the law of large numbers \cite{MCbook}, $\text{Fr}(C_{[0,T],N},M)$ converges with probability 1 towards $\Pr(C_{[0,T],N})$
\begin{equation}
\Pr\left(\lim_{M\rightarrow \infty}\text{Fr}(C_{[0,T],N},M)=\Pr(C_{[0,T],N})\right)=1
\end{equation}
therefore,
\begin{equation}
\Pr\left(\lim_{M,N\rightarrow \infty}\text{Fr}(C_{[0,T],N},M)=\Pr(C_{[0,T]})\right)=1.
\end{equation}
By the central limit theorem \cite{MCbook}, $\text{Fr}(C_{[0,T],N},M)$ is approximately normally distributed $\text{Fr}(C_{[0,T],N},M)\overset{\text{approx.}}{\sim}\mathcal{N}(\mathbb{E}\left[\text{Fr}(C_{[0,T],N},M)\right],\mathbb{D}\left[\text{Fr}(C_{[0,T],N},M)\right])$ with
\begin{align}
\mathbb{E}\left[\text{Fr}(C_{[0,T],N},M)\right]=&\Pr(C_{[0,T],N})\\
\mathbb{D}\left[\text{Fr}(C_{[0,T],N},M)\right]=&\frac{1}{M}\Pr(C_{[0,T],N})\left(1-\Pr(C_{[0,T],N}))\right).
\end{align}
Notice that
\begin{equation}
\Pr\left(\frac{\left|\Pr(C_{[0,T],N})-\text{Fr}(C_{[0,T],N},M)\right|}{\sqrt{\frac{1}{M}\Pr(C_{[0,T],N})\left(1-\Pr(C_{[0,T],N})\right)}}<z_{\alpha/2}\right)\approx \alpha
\end{equation}
and $\Pr(C_{[0,T]})\approx\Pr(C_{[0,T],N})\approx\text{Fr}(C_{[0,T],N},M)$, eventually, $\hat{\Pr}(C_{[0,T]})=\text{Fr}(C_{[0,T],N},M)$ is the MC estimation of $\Pr(C_{[0,T]})$ with approximate $100\alpha\%$ confidence in the estimation error
\begin{small}
\begin{equation}
\left|\Pr(C_{[0,T]})-\text{Fr}(C_{[0,T],N},M)\right|\le z_{\alpha/2}\sqrt{\frac{1}{M}\text{Fr}(C_{[0,T],N},M)\left(1-\text{Fr}(C_{[0,T],N},M)\right)}.
\end{equation}
\end{small}
$\square$
\end{document}